# An Efficient Evolutionary Based Method For Image Segmentation

Roohollah Aslanzadeh[1]
Kazem Qazanfari[2]
Mohammad Rahmati[1]
[1]Department of Computer Engineering and Information Technology, Amirkabir University of Technology
[2]Department of Computer Science, George Washington University

**Abstract**

The goal of this paper is to present a new efficient image segmentation method based on evolutionary computation. This paper starts by describing and introducing a model inspired from human behavior. Based on this model, a four layer process for image segmentation is proposed using the split/merge approach. In the first layer, an image is split into numerous regions using the watershed algorithm. In the second layer, a co-evolutionary process is applied to form centers of finals segments by merging similar primary regions. In the third layer, a meta-heuristic process uses two operators to connect the residual regions to their corresponding determined centers. In the final layer, an evolutionary algorithm is used to combine the resulted similar and neighbor regions. Different layers of the algorithm are totally independent, therefore for certain applications a specific layer can be changed without constraint of changing other layers.

Some properties of this algorithm like the flexibility of its method, the ability to use different feature vectors for segmentation (grayscale, color, texture, etc), the ability to control uniformity and the number of final segments using free parameters and also maintaining small regions, makes it possible to apply the algorithm to different applications. Moreover, the independence of each region from other regions in the second layer, and the independence of centers in the third layer, makes parallel implementation possible. As a result the algorithm speed will increase.

The presented algorithm was tested on a standard dataset (BSDS 300) of images, and the region boundaries were compared with different people segmentation contours. Results show the efficiency of the algorithm and its improvement to similar methods. As an instance, in 70% of tested images, results are better than ACT algorithm, besides in 100% of tested images, we had better results in comparison with VSP algorithm.

## 1. Introduction

Image segmentation is the process of partitioning a digital image into multiple segments. The goal of segmentation is to simplify and/or transform the representation of an image into something that is more meaningful and easier to analyze. Image segmentation is typically used to locate objects and boundaries (lines, curves, etc.) in images. The first studies on image segmentation were done forty years ago. Roberts's edge detector is one of the first edge detectors that proposed in 1963 [1]. This operator is the first progress in the field of image segmentation and after that, many segmentation techniques and algorithms have been proposed.

Segmentation algorithms are faced with many difficulties; Local structures of images such as flat regions and boundaries are appeared differently; image segmentation is a low level process in image analysis, so, segmentation algorithms partition images using low level concepts of images such as pixel intensity, pixel color, texture and etc. Another difficulty is the lack of accurate criteria for evaluating segmentation algorithms.

In this paper, a segmentation method is considered efficient if have these features: Continuity of contours [2], appropriate number of segments [2], free from any thresholding value [2], resistance against noise, low time complexity, usefulness presentation of results for higher level of image analysis, and finally each detected segment has to be homogenous and has uniform structure and adjacent segments have to have sharp difference (Color, intensity, texture, etc).

Satisfying above features causes universality of a segmentation method, so this segmentation method can be used in variety of applications. The contribution of this paper is to propose a novel evolution based segmentation algorithm that satisfies above features. This method is inspired from a natural behavior of human groups that are spread through a topographic plane and willing to have a social life. Proposed method includes multi independent levels, results in capability of altering each level separately to optimize the overall performance for specific application. These levels are: watershed algorithm, co-evolution process, immigration-deportation process, emerging process.

This paper is organized as follows: Section 2 briefly reviews some relative segmentation methods. Section 3 introduces the proposed model, and then based on this model, a novel image segmentation algorithm is proposed. Section 4 gives experimental results on the Berkeley segmentation database, and compares our method to other existing algorithms. Finally, section5 concludes the paper.

## 2. Background

Generally, there are two groups of segmentation algorithms [3]: 1) Boundary based methods which find objects' contour using dissociation feature and 2) Region based methods that detect objects using similarity features of pixels. Boundary based segmentation algorithms detect objects using dissociations in pixels. The primal methods of edge detection such as Roberts, Sobel, Canny and Perwitt are based on probability of edges in special spot based on some local measurements. Recently, some learning based segmentation methods are proposed which use the color, intensity and texture of image as suitable features for detecting segments [4-6]. For example, Martin and et al. [4] define some gradient operators on color components, intensity and texture, then by using these features and a logical regression classifier, they estimate edges strength. Also Dalar and et al. [5] proposed a supervised learning algorithm for edge and object boundary detection; the algorithm extracts and combines a large number of features across different scales in order to learn a discriminative model using an extended version of the Probabilistic Boosting Tree classification algorithm [6]. Wang and et al. [7] proposed an image segmentation method which considers the local image information by describing it as a novel local signed difference (LSD) energy, which possesses both local separability and global consistency. In another study, Zengwei and et al. [8] proposed an image segmentation algorithm based on the adaptive edge detection and an improved mean shift is proposed. In their method, an adaptive threshold algorithm is applied to improve Canny operator in edge detection. Also, Tan and et al. [9] presented a novel histogram thresholding – fuzzy C-means hybrid (HTFCM) approach for image segmentation which first applies a histogram thresholding technique to obtain all possible uniform regions in the color image and then, a Fuzzy C-means (FCM) algorithm is utilized to improve the compactness of the clusters forming these uniform regions.

The goal of region based segmentation algorithms is to detect and recognize regions that satisfy defined homogeneity criteria [10, 11]. Splitting-Merging based methods are kind of these methods [12]. These methods have two steps. In the first step, image is split recursively while satisfying homogeneity criteria and in the second step some regions are merged if the produced region satisfying homogeneity criteria.

Flez-Hutt method is a graph based technique for emerging image regions proposed by Felzenszwalb [13]. Graph nodes are image pixels and their edges weight is dissimilarity value between them. In each iteration two nodes $C_1$ and $C_2$ is merged together if their edge weight is lower than a particular value. In another work, Shi and Malik [14] proposed a graph based segmentation methods which rather than focusing on local features and their uniformity in the image data, focus at selecting the global impression of an image. They treated image segmentation as a graph partitioning problem and suggest a global criterion, the normalized cut, for partitioning the graph. Also, Freedman and Daniel [15] proposed two improvements to the image graph used by the Random Walker method. First, they proposed a new way of computing the edge weights. The second improvement, the traditional graph has a vertex set which is the set of pixels and edges between each pair of neighboring pixels. They substitute a smaller, irregular graph based on Mean Shift over segmentation.

Also Cour and et al. [16] present a multi scale spectral image segmentation algorithm. In contrast to most multi scale image processing, this algorithm works on multiple scales of the image in parallel, without iteration, to get both coarse and fine level details. The algorithm is computationally efficient, allowing to segment large images. Also, Nguyen and et al. [17] proposed a robust and accurate interactive method based on the recently developed continuous-domain convex active contour model.

Duarte and et al. [18] proposed a splitting-merging based segmentation method that uses a heuristic technique. The splitting step is done by watershed algorithm. The merging step uses a weighted undirected graph, called MRAG to join two similar regions. The segments resulted by the first step are the graph's nodes and their edge weight is dissimilarity value between these nodes. Finally, the outcome graph is iteratively partitioned in a hierarchical way into two sub graphs, corresponding to the two most significant components of the actual image, until a termination condition is met. This graph-partitioning task is solved by a variation of the min-cut problem (normalized cut) using a hierarchical social (HS) Meta heuristic

Mean Shift method [19] proposed by Comaniciu and Meer is a general nonparametric technique for analysis of a complex multimodal feature space and to form arbitrarily shaped clusters in it. The essential computational part of the technique is an old pattern recognition process, the mean shift. In another work, Arbelaez and et al. [20] addressed the problem of segmenting and recognizing objects in real world images, focusing on challenging articulated categories such as humans and other animals. For this purpose, they proposed a novel design for region-based object detectors that integrates efficiently top-down information from scanning-windows part models and global appearance cues.

Gpb-owt-ucm method proposed by Arbeláez and et al. [21] results a hierarchical segmentation for each edge detection method. They proposed a new version of watershed algorithm, called Oriented Watershed Transform (OWT) to form initial regions from contours, followed by construction of an Ultra metric Contour Map (UCM) [22] defining a hierarchical segmentation. Also using edge detection method gPb [23] as the input of this method (i.e. gPb-owt-ucm) causes to increase the performance of this method. Also, Gupta and et al. [24] address the problems of contour detection, bottom up grouping and semantic segmentation using RGB-D data. They proposed algorithms for object boundary detection and hierarchical segmentation that generalize the gPb − ucm approach of [22] by making effective use of depth information.

Population based segmentation methods inspired from behavior of population and simulate their intelligence to solve the segmentation problem. Some of well known techniques in the field of image segmentation are PSO, Ant colony, bee colony and etc.

Deriving from the artificial life theory, Chen and et al. [25] proposed an artificial co-evolving tribes model and applies it to solve the image segmentation problem. During the evolution process, the individuals in this model making up the tribes effect communication cooperatively from one agent to the other in order to increase the homogeneity of the ensemble of the image regions. Two remarkable properties, that is, the monotone contraction and the conservation of the system are proved. Guo and Li [26] proposed a PSO based segmentation algorithm using multilevel thresholding method. This method resolves the problem of complexity time by maximization entropy, but the difficulty of this method is determination of threshold levels before performing algorithm.

Huang and et al. [27] presented a segmentation algorithm based on artificial ant colonies (AC). In this model, each ant can memorize a reference object, which will be refreshed when it finds a new target. A fuzzy connectedness measure is adopted to evaluate the similarity between target and the reference object. The behavior of an ant is affected by the neighbors and the cooperation between ants is performed by exchanging information through pheromone updating. Also Han and Shi [28] proposed a fuzzy ant colony algorithm (ACA), inspired by the food-searching behavior of ants, for image segmentation. Three features such as gray value, gradient and neighborhood of the pixels, are extracted for the searching and clustering process. This method has been improved by initializing the clustering centers and enhancing the heuristic function to accelerate the searching process.

Pablo and et al. [29] investigated two fundamental problems in computer vision: contour detection and image segmentation. Their contour detector combines multiple local cues into a globalization framework based on spectral clustering and their segmentation algorithm consists of generic machinery for transforming the output of any contour detector into a hierarchical region tree. Also, Hariharan and et al. [30] studied the challenging problem of localizing and classifying category-specific object contours in real world images. For this purpose, they presented a simple yet effective method for combining generic object detectors with bottom up contours to identify object contours.

Melkemi and et al. [31] proposed a distributed image segmentation algorithm structured as a multi agent system composed of a set of segmentation agents and a coordinator agent. Each segmentation agent performs the iterated conditional modes method, known as ICM, in applications based on Markov random fields, to obtain a sub-optimal segmented image. This method is robust against noise but has high complexity time. Also Lai and chang [32] proposed a clustering based segmentation method using genetic algorithm. This method uses a hierarchical structure of chromosomes that classified image automatically into appropriate segments. This method has been applied on medical images and without any information about number of clusters. Also this method is robust against noise. Also, Yang and et al. [33] formulated a layered model for object detection and image segmentation. They describe a generative probabilistic model that composites the output of a bank of object detectors in order to define shape masks and explain the appearance, depth ordering, and labels of all pixels in an image.

3. **The proposed model and algorithm**
In this paper a novel image segmentation method is proposed that is based on the evolutionary computing algorithms. The proposed method is independent from the image content; so this method can be used in the segmentation stage of verity applications. Also this method includes multi independent steps, results in capability of altering each step separately to optimize the overall performance for specific application.
In this section, first, an intuitive model which inspired from the nature is proposed. This model presents the behavior of human groups that are spread through a topographic plane and willing to have a social life. Then this model mapped out to the image space and a multistep segmentation algorithm is proposed for image segmentation.
3.1. **The Proposed model**
In this section, a new model is introduced that is based on the human group's behavior to form different tribes and to determine their zones. Suppose that these human groups are spread throughout a topographical plane. Based on our model, these groups establish some tribes so that the groups of each tribe have maximum similarity in their properties and near to each other locally. In this paper, each tribe represents an image segments, each tribe's people is stand for a segment's pixel and finally, each group is stand for a primitive segment that generated by the watershed algorithm in the first step of our algorithm.

The procedure of formation the tribe in the proposed model is presented in following. This procedure includes following steps:

1) In the beginning of process, the people which located in a flat region of plane set up a group. There are not any tall mountains between group's people and these people have similar features.
2) Each group makes relation with its neighbor groups. Then based on the relation strength between two neighbor groups, each group comes near or keeps out from other group and then gets effect from it.
    a. In the beginning of this step, the relation strength of two neighbor groups is depended on the obstacles and difficulties between them.
    b. Each group attempts to make a relation with other groups and tries to pass obstacles and difficulties between them. Then, the relation strength between two groups determined based on their distance and similarity of their properties.
    c. Each group investigates its environment features. If this group has maximum similarity with its neighborhood groups, then this group has been matured and inhabits in its location.
    d. If a set of matured groups 1) have strong relations, 2) located in the same neighborhood and 3) number of groups is enough to set up a tribe; then these groups united together and create an initial tribe.
3) Following steps get start after finishing previous step:
    a. A group that is not member of a tribe, attempts to migrate to one of its neighborhood tribes. This group selects a tribe based on the distance and similarity of itself to the tribe's groups.
    b. To keep tribes integrity, each tribe deports the immigrated group which is not similar to initial tribe's groups. The deported group has to migrate to another tribe.
4) At the end of step 3, the groups of each tribe have maximum similarity but, maybe there are some neighbor tribes that do not have remarkable different. So, to increase the resultant efficiency, those tribes make a bigger tribe, united together and remove their boundaries.
5) At the end of step 4, there are some tribes that each one has maximum dissimilarity compare to other tribes and its groups have maximum match. This result is closely like to our desired in the field of image segmentation.

In the next section, a novel segmentation algorithm is proposed that is inspired from the mentioned model. The formation process of segments in this algorithm is similar to the formation procedure for tribes. Also each final segment has similar features to a final tribe in mentioned model.

### 3.2.  The Proposed algorithm

In this section, a new image segmentation algorithm is presented that is based on the proposed model in the previous section. Similar to that model, proposed algorithm has 4 main steps. Figure 1 shows these steps of our algorithm. To distinguish between equivalents of the group and tribe in the proposed segmentation method, the phrases "primitive segment" and "segment" are used as equivalent of "group" and "tribe" words respectively. The remaining of this section, describes each step of this algorithm.

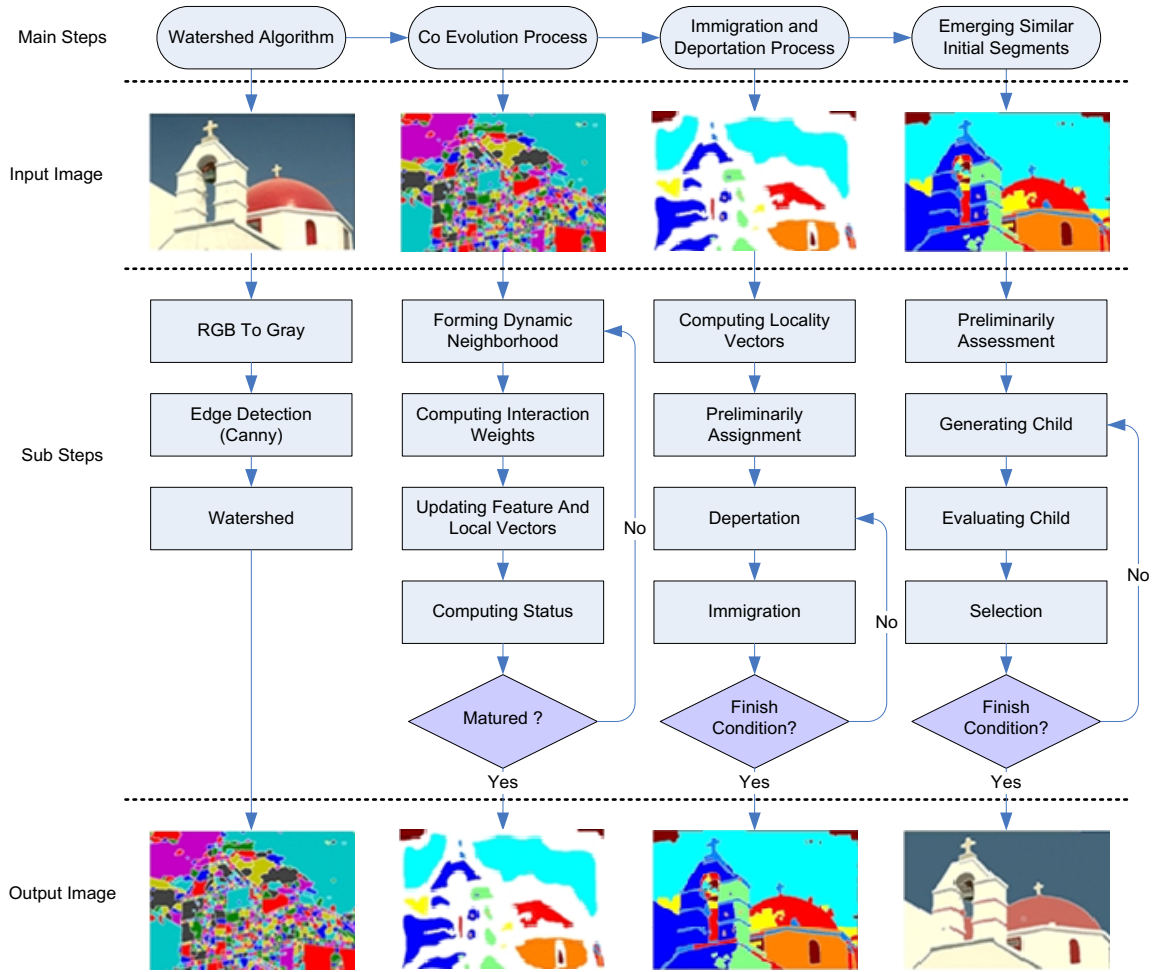

Figure 1. A brief overview of proposed method (Main steps and sub steps).

### 3.2.1. Step 1: Watershed algorithm

The first step of proposed model is simulated by watershed algorithm. The input of this step is a gray scale image. So, if the main input image is RGB, then a gray scale copy is used. The outcome of this step is a segmented image that has a lot of primitive segments. Each primitive segment represents a group of people in the proposed model.

Watershed algorithm has two important advantages including: a) low time complexity and b) Continuous boundary of primitive segments. Either main image or its edge could be used as the input of watershed algorithm, but the experimental results show that this algorithm results more primitive segments in the way that main image is used. So, first, our method uses an edge algorithm (canny) to compute the probability matrix of edges for the gray scale image. Then, the watershed algorithm is used with the probability matrix of edge as watershed algorithm's input. Figure 2 shows the results of primitive segmentation by the watershed algorithm on the probability matrix of edge (canny) and main image.

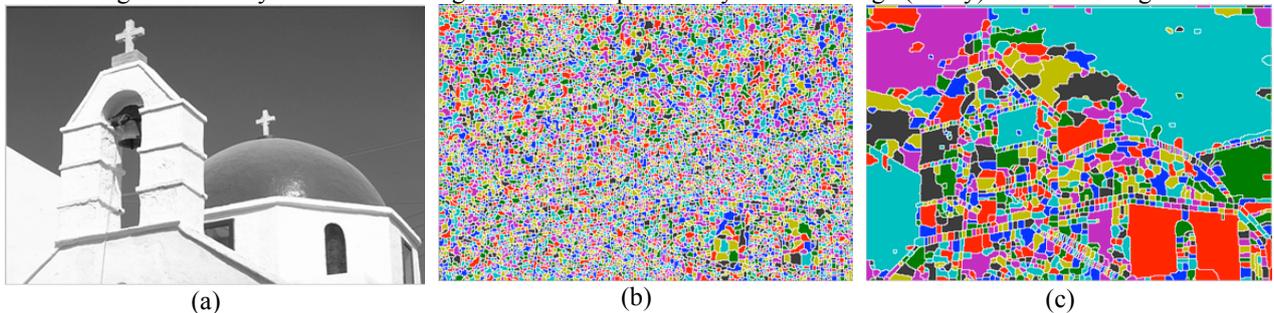

(a)          (b)          (c)

Figure 2.The results of primitive segmentation by watershed algorithm. Input gray scale image (a). Primitive segments when the gray scale image is used as input (b) and when the probability matrix of edge is used as input (c). The numbers of primitive segments are 9194 and 1165 for (b) and (c) respectively.

### 3.2.2. Step 2: Co evolution Process

This step of proposed algorithm simulates the second step of described model. This step classifies primitive segments into two classes: matured and immature. At the end of this step, each primitive matured segment may be connected to some similar primitive matured segments in their neighborhood. These connected primitive segments labeled with the same name and form an initial segment (equivalent to a tribe in the model). Also immature segments do not have any label. That is because of they do not qualify the matured conditions or are not big enough to form a bigger segment.
In this step, a primitive segment is considered as a group of people, and the boundaries between primitive segments (resulted by watershed algorithm) are considered as obstacles and mountain in the proposed model. Each primitive segment has two feature vectors: 1) $s_i^{(t)}$: A four dimensional spatial vector including length, width, and center (x,y) of surrounding rectangle of primitive segment i. 2) $p_i^{(t)}$: A One (for gray scale images) or three (for color images) dimensional color vector including the average of primitive segment's pixels color.
Figure 1 shows the sub steps of co evolution process for each primitive segment. This process is kept on for each primitive segment separately, so, the total process of primitive segments would be co evolution. In the next subsections, we describe each sub steps of this co evolution process.

### 3.2.2.1. Defining Dynamic Neighborhood
The dynamic neighborhood of each primitive segment *i* in iteration *t* is computed by equation (1):
$$\mathcal{N}_r^{(t)}(i) = \{j | d(s_i^{(t)}, s_j^{(t)}) \leq r\} \tag{1}$$
Where, $d(s_i^{(t)}, s_j^{(t)})$ is the spatial distance of two primitive segment *i* and *j* and computed by equation (2):
$$d(s_i^{(t)}, s_j^{(t)}) = max(dx(s_i^{(t)}, s_j^{(t)}), 0) + max(dy(s_i^{(t)}, s_j^{(t)}), 0) \tag{2}$$
$$dx(s_i^{(t)}, s_j^{(t)}) = |s_i^{(t)}(1) - s_j^{(t)}(1)| - (s_i^{(t)}(2) + s_j^{(t)}(2)) \tag{3}$$
$$dy(s_i^t, s_j^t) = |s_i^{(t)}(3) - s_j^{(t)}(3)| - (s_i^{(t)}(4) + s_j^{(t)}(4)) \tag{4}$$
Where, $s^{(t)}(1)$ and $s^{(t)}(3)$, are the coordinate (x,y) of surrounding rectangle's center of a primitive segment and $s^{(t)}(2)$ and $s^{(t)}(4)$ are the half length and half width of this rectangle. Figure 3 shows *dx* and *dy* graphically for two typical primitive segments.

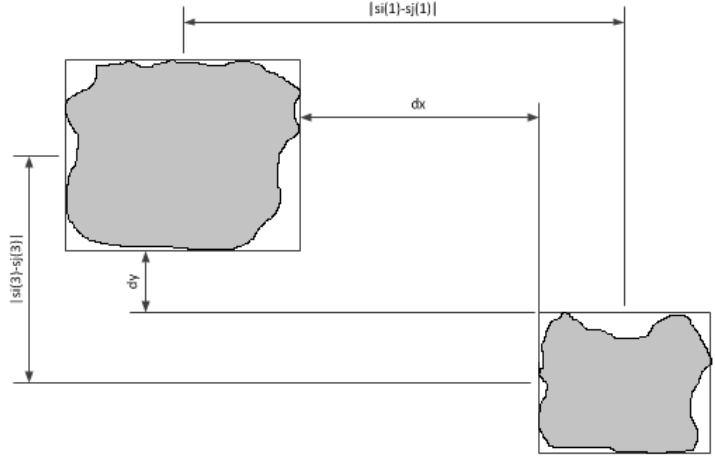

Figure 3. Spatial distance *dx* and *dy* between two primitive segments.

### 3.2.2.2. Defining Interaction Weights and Updating Rules
Each primitive segment *i* makes relation with each its neighbor primitive segment *j* in each iteration *t*. The strength of this relation is depended on the spatial distance and the similarity of these two primitive segments and could be computed by equation (5).
$$w_{i,j}^{(t)} = F(d(s_i^{(t)}, s_j^{(t)}), \|p_i^{(t)} - p_j^{(t)}\|) \tag{5}$$
Interaction weights have following features:
1) $w_{i,j}^{(t)} = w_{j,i}^{(t)}$.
2) Consider two primitive segments *i* and *j* located in the same initial segment and primitive segment *k* in another initial segment. It's expected that $w_{i,j}^{(t)} \gg w_{i,k}^{(t)}$.
3) If two primitive segment *i* and *j* have same feature vector (spatial and color), then their weight vector $w_{i,j}^{(t)}$ has maximum value.

In the first iteration of this step, the interaction weight of two primitive segments is depended on the boundary between these two segments. Also based on Ref. [34] the maximum energy of this boundary is considered as the difficulty of passing this boundary for $\mathcal{E}^*$. Therefore, the initial value of interaction weights computed by equation (6):
$$w_{i,j}^{(0)} = \begin{cases} 1 - max_{D_k \in L_{i,j}}(1 - \exp(-\mathcal{E}^*(D_k)/\sigma_\omega)) & \|p_i^{(t)} - p_j^{(t)}\| \leq \theta_p \\ 0 & otherwise \end{cases} \tag{6}$$

Where, $L_{i,j}$ is the set of boundaries between two primitive segments $i, j$ and $\sigma_\omega$ normalizes $\mathcal{E}^*$ and based on Ref. [34] is equal to 0.02.

In the other iteration, $(t > 1)$, primitive segments can pass more difficult obstacles and boundaries. In these iterations, the interaction weight between two primitive segments is depended on locality of them (spatial vectors) and similarity of their color vectors and can be computed by equation (7):

$$w_{i,j}^{(t)} = \begin{cases} 1 - \left(d(s_i^{(t)}, s_j^{(t)})/2r + \|p_i^{(t)} - p_j^{(t)}\|/2\theta_p\right) & \|p_i^{(t)} - p_j^{(t)}\| \leq \theta_p \\ 0 & otherwise \end{cases} \quad (7)$$

As shown in equation (7), interaction weight $w_{i,j}^{(t)}$ has all three mentioned features and $w_{i,j}^{(t)} \in [0,1]$.

### 3.2.2.3. Updating Feature Vectors

After computing the interaction weights, each primitive segment updates its spatial vectors as following:

$$s_i^{(t+1)}(1) = s_i^{(t)}(1) + \Delta s_i(1) \quad (8)$$
$$s_i^{(t+1)}(3) = s_i^{(t)}(3) + \Delta s_i(3) \quad (9)$$
$$s_i^{(t+1)}(2) = s_i^{(t)}(2) * 0.98 \quad (10)$$
$$s_i^{(t+1)}(4) = s_i^{(t)}(4) * 0.98 \quad (11)$$

Where, $\Delta s_i$ is the movement direction of primitive segment $i$ and could be computed by equation (12):

$$\Delta s_i(x) = \frac{1}{\mathcal{M}_i^{(t)}} \sum_{j \in \mathcal{N}^{(t)}(i)} w_{i,j}^{(t)} \left(s_j^{(t)}(x) - s_i^{(t)}(x)\right) \quad x = 1,3 \quad (12)$$

Where, $\mathcal{M}_i^{(t)} = \sum_{j \in \mathcal{N}^{(t)}(i)} \sup(w_{i,j}^{(t)})$ and normalizes the total interaction weights. These equations show that each primitive segment comes near to similar neighbors and its dimensions are came smaller (or the people of group are centralized).

While moving to location $s_i^{(t+1)}$ each primitive segment updates the color vector based on its new environment using equation (13):

$$p_i^{(t+1)} = \frac{1}{\mathcal{M}_i^{(t)}} \sum_{j \in \mathcal{N}^{(t)}(i)} w_{i,j}^{(t)} p_j^{(t)} + \lambda_i^{(t)} p_i^{(t)} \quad (13)$$

Parameter $\lambda_i^{(t)}$ is the status pointer of the primitive segment $i$ and shows the adaption strength of this primitive segment in its new environment. The value of this parameter is between [0, 1] and as the adaption is more strong, this parameter has lower value. This parameter computed by following equation (14):

$$\lambda_i^{(t)} = 1 - \frac{1}{\mathcal{M}_i^{(t)}} \sum_{j \in \mathcal{N}^{(t)}(i)} w_{i,j}^{(t)} \quad (14)$$

If $\lambda_U \leq \lambda_i^{(t)} \leq 1$, then this primitive segment does not have strong adaptation with its new environment, so, the features vectors are updated as following:

$$p_i^{(t)} = p_j^{(t)} \quad s_i^{(t)} = s_j^{(t)} \quad (15)$$

Where, $j = \arg\min_j \{d(s_i^{(t)}, s_j^{(t)})\}$ and $\lambda_i^{(t)} < \lambda_U$. This strategy removes the noisy primitive segments (that usually have little pixels) and the small dissociations of image.

Else if $0 \leq \lambda_i^{(t)} \leq \lambda_L$, then primitive segment $i$ has adequate energy to be matured. In this situation, two vectors $s_i^{(t)}$ and $p_i^{(t)}$ is never changed and this primitive segment is disable and its co-evolution process is stopped. In this paper $\lambda_U$ and $\lambda_L$ are equal to 0.98 and 0.02 respectively.

### 3.2.2.4. Construction of Initial Zones

The co-evolution process of primitive segments is stopped if the number of matured primitive segments is not changed after n iteration. In our experiment, we consider n=20. Then, the initial segments are formed as following:

*Each matured primitive segment could be as the center of its initial segment. This primitive segment i is connected to each its neighbor matured primitive segment j if $w_{i,j}^{(t)} \geq \xi_c$. In the proposed algorithm, $\xi_c$ is equal to 0.6. If the number of pixels that are located in a set of connected primitive segments be more than $\delta_t$ percent of total image pixels, then this set forms an initial segment with label $T_k$ (also all primitive segments in the initial segment $T_k$ labeled as $T_k$ and form the **core** of initial segment). Parameter $\delta_t$ controls the minimum size of an initial segment and in this algorithm is equal to 0.03. So, other sets (that include some matured primitive segments) which are smaller than $\delta_t$ and the immature primitive segments cannot form any initial segment. All these primitive segments are known as unlabeled primitive segments. These primitive segments connect to the fittest initial segment in the next step of our method.*

### 3.2.3. Immigration and Deportation Processes

The unlabeled primitive segments are significant segments, since most of them are corresponding to the boundary primitive segments in the structure of the final segments. Therefore, this step of the proposed algorithm has important role in the quality of final segments.

In this step, a meta-heuristic evolution algorithm is proposed. The sub steps of this process are shown as a block diagram in figure 1. As shown in this diagram, in the beginning of this step, each initial segment $T_k$ (refer to **core** of this segment) has a spatial vector $C_{T_k}$ that is computed by following Eq. (16) and (17):

$$C_{T_k}(1) = \sum_{primitive\ segment\ i \in core\ T_k} s_i(1) / Number\ of\ primitive\ segments\ located\ in\ core\ T_k \tag{16}$$

$$C_{T_k}(2) = \sum_{primitive\ segment\ i \in core\ T_k} s_i(3) / Number\ of\ primitive\ segments\ located\ in\ core\ T_k \tag{17}$$

### 3.2.3.1. Preliminary Assignment of Unlabeled Primitive Segment to an Initial Segment

The second step of proposed meta-heuristic method is to assign each unlabeled primitive segment to an initial segment. In our method, each unlabeled primitive segment migrates to the nearest initial segment. This assignment reduces the number of iterations in the deportation-immigration process. After this assignment, the process of deportation-immigration continues until the stop condition is satisfied. Also, in each iteration, the process of deportation-immigration is applied on all unlabeled primitive segments that are not member of segments cores parallel.

### 3.2.3.2. Deportation Process

The deportation operator computes the deportation probability of each migrated primitive segment using the evaluation function F. For each migrated primitive segment, a random number between [0 1] is generated, if deportation probability is more than this number then the migrated primitive segment is deported and its label is changed to 0. Deportation probability of the primitive segment $i$ that migrated to the initial segment $T_k$ computed by following equation:

$$P_{deport}(i) = \frac{F(i) - \min_{j \in T_k} F(j)}{\max_{j \in T_k} F(j)} \tag{18}$$

Where, $js$ are the primitive segments which are not member of initial segment core $T_k$. Also, the evaluation function F computes the maximum contrast between a migrated primitive segment $i$ and each primitive segment $l$ which is member of initial core $T_k$ using equation (19):

$$F(i) = \max\{contrast(i, l), l \in T_k\} \tag{19}$$

Where, $l$ is a primitive segment that is the member of initial segment core $T_k$. Also $contrast(i, l)$ is the Manhattan distance between two color vectors of primitive segments $i$ and $l$.

### 3.2.3.3. Immigration Process

The immigration process is applied only on deported primitive segments. The immigration operator computes the membership worth through all initial segments for each deported primitive segment by evaluation function G. For each deported primitive segment, the fittest initial segment is selected. Then the deported primitive segment migrates to this initial segment.

The membership worth of primitive segment $i$ for initial segment $T_k$ is computed by evaluation function G:

$$G(i, T_k) = \alpha \frac{G_1(i, T_k)}{\max G_1} + (\alpha - 1) \frac{G_2(i, T_k)}{\max G_2} \tag{20}$$

$$G_1(i, T_k) = \max\{contrast(i, l), l \in T_k\} \tag{21}$$

$$G_2(i, T_k) = d(s_i, C_{T_k}) \tag{22}$$

Function $G_1(i, T_k)$ computes the similarity of color features similar to equation (19) and function $G_2(i, T_k)$ computes the closeness of the primitive segment $i$ to the initial segment $T_k$. $C_{T_k}$ is computed using equation (16) and (17). Also, $\alpha$ is the importance coefficient of color or spatial features and it is between [0 1]. As $\alpha$ is higher, the color similarity would be more important, otherwise the spatial closeness would be more important. In our method, $\alpha$ is equal to 0.9. As realized from the definition of G, for deported primitive segment $i$ and initial segment $T_k$, as G is lower, this initial segment is more fit for this primitive segment.

### 3.2.3.4. Stop Conditions

The process of deportation-immigration is continued until; 1) The number of iterations reaches to m **or** 2) Deportation-immigration rate reaches under r. After stopping deportation-immigration process, all primitive segments have label and each one is member one initial segment. Also, whole image is covered by all initial segments. At the end of this step, maybe there are two or more similar initial segments that are separated because of the unlabeled primitive segments resulted by the second step of proposed method. Now, after connecting these unlabeled primitive segments to those initial segments (step 3), to achieve an efficient segmentation, these similar initial segments have to be merged together. In the final step of our algorithm, an iterative process is proposed to reach this goal.

### 3.2.4. Emerging Similar Initial Segments

The forth step of our model is simulated by the initial segments emerging process. In this step, the boundaries between two similar neighbor initial segments are removed by an iterative process. The result of each iteration is compared with the result of previous iteration. Based on an evaluation criterion, if the new iteration shows better efficiency than the previous iteration, the previous segments are replaced by new segments. The emerging method that is used in this step is a kind of genetic algorithms. The introduced genetic algorithm has an intelligent mutation operator that increases the rate of convergence and the efficiency of segmentation. The sub steps of this process are shown as a block diagram in Figure 1.

The population size of proposed genetic algorithm is one. So, in the each iteration, this algorithm generates only a child by the intelligent mutation operator, then, in the selection step, selects the fittest one among child and parent as the next generation based on the evaluation criterion. Therefore, this algorithm is called genetic algorithm (1, 1+1).

#### 3.2.4.1. Chromosome Coding Or Representation

In the proposed method, a binary vector is used as a genetic representation similar to Ref. [2]. Suppose that the results of previous step are R initial segments $(r_1, r_2, ..., r_R)$ and N edges $(e_1, e_2, ..., e_N)$. So, the chromosome is a binary string of $a_1 a_2 ... a_N$. Where, $a_i = 1$ means preserving edge $e_i$ and $a_i = 0$ means removing this edge. Figure 4 shows an example of the coding process for this step.

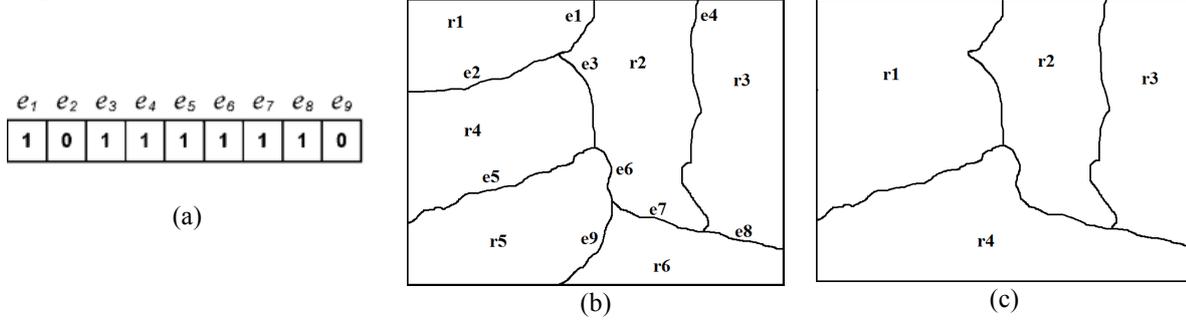

Figure 4. An example of the coding process for emerging step. (a): A chromosome, (b) The segmented image (result of step 3). (c): The new segmented image after one iteration based on (a).

#### 3.2.4.2. Evaluation Function

To evaluate the image segments of new iteration, equations (23) and (24) are used:

$$\overline{D}(I) = \frac{1}{N} \sum_{i=1}^{N} \frac{S_i}{S_I} \overline{D}(R_i) \quad (23)$$

$$D(R_i, R_j) = \frac{|m_{gl}(R_i) - m_{gl}(R_j)|}{NG} \quad (24)$$

These equations are inspired from an unsupervised evaluation method $F_{RC}$ [35]. This method finds a chromosome that has higher $\overline{D}(I)$. Higher $\overline{D}(I)$ expresses lower different between two initial segments. Zhang and et al. [36] evaluated 9 unsupervised methods and showed that $F_{RC}$ method is one of the better methods which could improve the efficiency of segmentation approaches that are similar to the human segmentation. Also, since similar segmentation to human is one of our considerations for an efficient segmentation algorithm, this method could be a profitable choice to reach that segmentation method.

Also $m_{gl}(R_x)$ is the intensity average of pixels which located in the initial segment $R_x$ and *NG* is the number of gray levels, e.g. is equal to 256 for the 8 bit depth images.

#### 3.2.4.3. Mutation Function

Mutation operator resets to zero each of chromosome bit $a_k$ (only $a_k = 1$, k =1, 2, …, N) with the probability of $P_{ij}$, where $a_k$ shows the existence of boundary between two initial segment i and j and computed as following:

$$\rho_{ij} = \begin{cases} 1 - \frac{|m_{gl}(R_i) - m_{gl}(R_j)|}{\theta_p} & |m_{gl}(R_i) - m_{gl}(R_j)| \leq \theta_p \\ 0 & otherwise \end{cases} \quad (25)$$

As mentioned before, $m_{gl}(R_x)$ is the intensity average of pixels which located in the initial segment $R_x$ and $\theta_p$ controls the maximum difference of two initial segments $R_i$ and $R_j$ to emerge together. The value of this parameter is considered equal to parameter $\theta_p$ in the second step of proposed method (Equations 6 and 7).

#### 3.2.4.4. Selection Method

The selection method is based on eligibility of chromosome. On the other word, the fittest chromosome is selected among child and its parent as the next generation based on the evaluation criterion.

#### 3.2.4.5. Stop Condition

After the special number of iterations (In the experiments, it's equal to 10), step 4 is stopped and the final chromosome expresses the state of boundaries between initial segments. If the chromosome bit $a_k$ is zero, then the edge $e_k$ is removed and two initial segments which separated by this edge are merged together. After the final step of our algorithm, the final segments are formed.

### 4. Results and Discussion

As mentioned before, similar segmentation to human is one of our considerations for an efficient segmentation algorithm, therefore, to evaluate the efficiency of our method the Berkeley Database BSDS300 [37] is used. BSDS300 has 300 images, which one segmented by several people. Also we use a boundary based evaluation technique proposed by martin et al. [4] that developed on BSDS300 to compute the accuracy of proposed method. This technique computes the precision, recall and F measure using equations (26), (27) and (28):

$$P = \frac{TP}{TP + FP} \tag{26}$$

$$R = \frac{TP}{TP + FN} \tag{27}$$

$$F = \frac{1}{\chi_0 * \frac{1}{P} + (1 - \chi_0) * \frac{1}{R}} \tag{28}$$

Where, precision is the probability that a machine-generated boundary pixel is a true boundary pixel. Precision is a measure of how much noise is in the output of the detector. Recall is the probability that a true boundary pixel is detected. Recall is a measure of how much of the ground truth is detected. Finally F is a measurement which is the harmonic mean of precision and recall.

### 4.1. Discussion on r, $\theta_p$ and Input Image type

The number of segments affects the performance of higher level processes of image analysis. The proposed algorithm has two parameters $r$ and $\theta_p$ to control the number of segments and therefore to improve the performance of specific application. Parameter r controls the neighborhood size, and $\theta_p$ determines the maximum acceptable difference between two adjacent primitive segments for locating in the same segment. It is clear that as the neighborhood size is increased, each primitive segment makes relation with farther primitive segment and if they are similar, they are placed in a same final segment, so, the number of final segments is decreased. Also, as $\theta_p$ is increased, the maximum acceptable difference between two primitive segments for locating in the same final segment is increased, so, the number of final segment is decreased. Therefore, different value for these parameter, result in different segments and different accuracy (F-measure). For example, table 1 shows the F-measure for two BSDS300 images (42049 and 216081) with different values for r and $\theta_p$.

Table 1 F-measure for two BSD300 images with different values for r and $\theta_p$.

| Image Code: 216081 | | | | | |
|---|---|---|---|---|---|
| $(r, \theta_p)$ | (9 , 17) | (10, 19) | (12 , 12) | (14 , 16) | (13 , 12) |
| F-measure | 0.84 | 0.83 | 0.80 | 0.78 | 0.70 |
| Image Code: 42049 | | | | | |
| $(r, \theta_p)$ | (11 , 14) | (14 , 16) | (10 , 19) | (9 , 17) | ( 12 , 12) |
| F-measure | 0.85 | 0.83 | 0.80 | 0.78 | 0.76 |

After processing all BSD300 images (gray and color), we conclude that for almost two-thirds of images, our method results better proficiency (F-measure) for colored version of images. Figure 5 shows the resulted F-measures by our method on 10 random selected images (gray and color) from BSD300. These F measures are the best value among variety values for two parameters r and $\theta_p$.

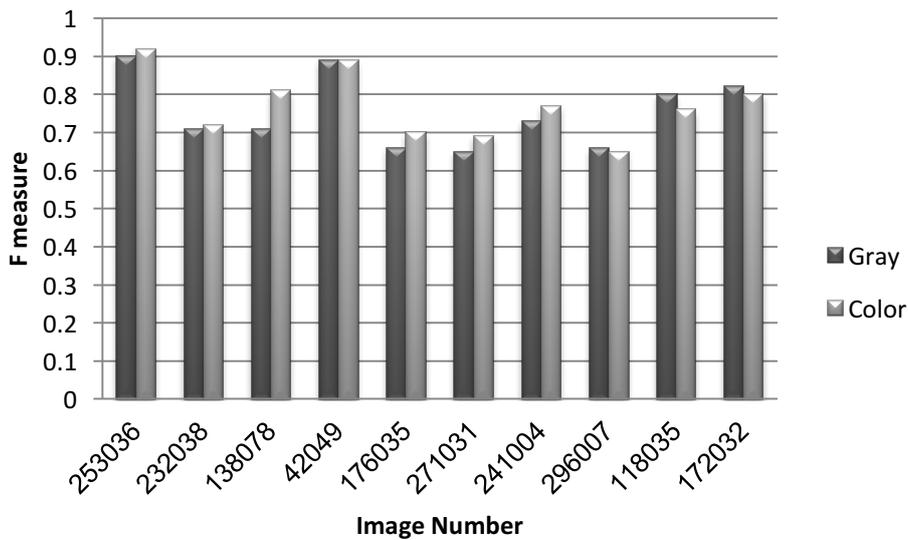

Figure 5. Proficiency of proposed method for ten color or gray scale image from BSD300. These F measures are the best value among variety values for two parameters r and $\theta_p$.

## 4.2. Comparison with other methods

In this section, we compare our method with some similar and recently methods. The first method is ACT method [25] which uses similar co-evolution algorithm as the second step of the proposed method. This method reported its result on 10 different images of BSD300. Figure 6 shows the F- measure for these segmented images produced by ACT and our method. As shown in this figure, our proposed algorithm results in higher F-measure in despite of some remarkable notices; a) ACT method compares the result segments with all different human produced segments and selects the fittest human produced segments for comparing, results in selecting the best F-measure. B) The authors of ACT method reported the results of their method only on ten images of BSD300; therefore, for comparison of our method against this method, we had to use only these ten images. C) Based on visual comparing on same images, proposed method can detect more conceptual small segments than ACT method, While, ACT method merges these segments with other segments. D) The time complexity of proposed method is very lower than ACT method. The time complexity of ACT and our method is $O(n^2)$ and $O(m^2)$ respectively where n is the number of image pixels and m is the number of primitive segment produced by watershed algorithm, so $m \ll n$.

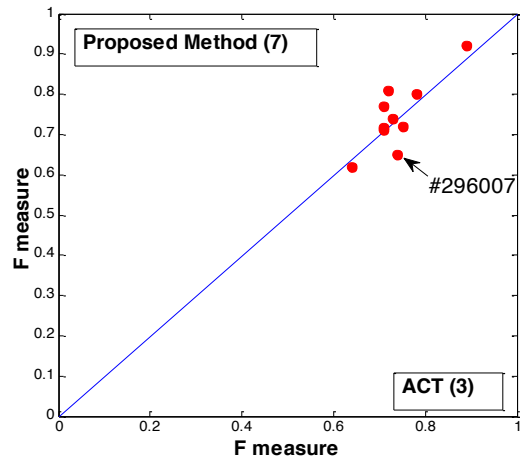

Figure 6 Comparison of F- measures for same segmented images (118035, 138078, 161062, 176035, 232038, 241004, 253036, 271031, 286092, and 296007) produced our method and ACT method [25].

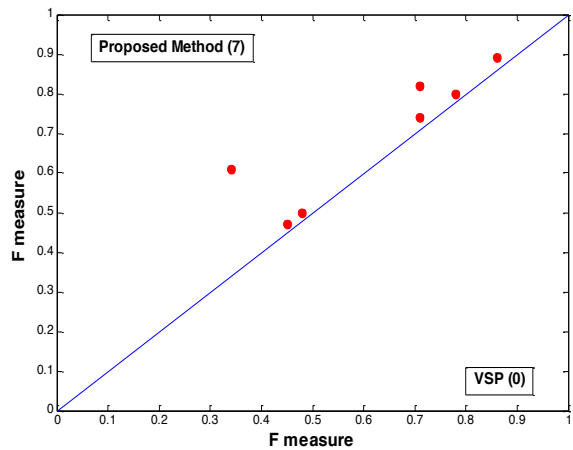

Figure 7 Comparison of F- measures for same segmented images (100080, 112082, 304034, 97017, 118035, 172032, 420049) produced our method and VSP method [38]

VSP method is a segmentation method that proposed by [38]. Similar to ACT methods, the authors of VSP method reported the results of their method only on seven images of BSD300; therefore, for comparison of our method against this method, we used these seven images. Figure 7 shows the F- measure for these segmented images produced by VSP and our method. As shown in this figure, our proposed algorithm results in higher F-measure.

Another similar segmentation method to our algorithm is HS algorithm [18]. This method is based on dividing and merging approach and used the watershed algorithm for dividing and a meta-heuristic method for merging processes. We compare our method with this method on only one common image from BSD300 (code 62096). Figure 8 shows the 4 outcome segments of proposed method and HS method. As shown in this figure, proposed method detects the sky as a one segment, while HS method distinguishes the same object in two separate segments. This phenomenon is occurred for the sea region too. Also this method uses MRAG matrix that occupies large amount of memory when the image size is high.

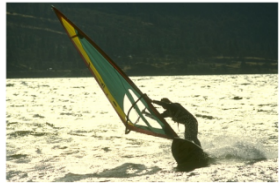

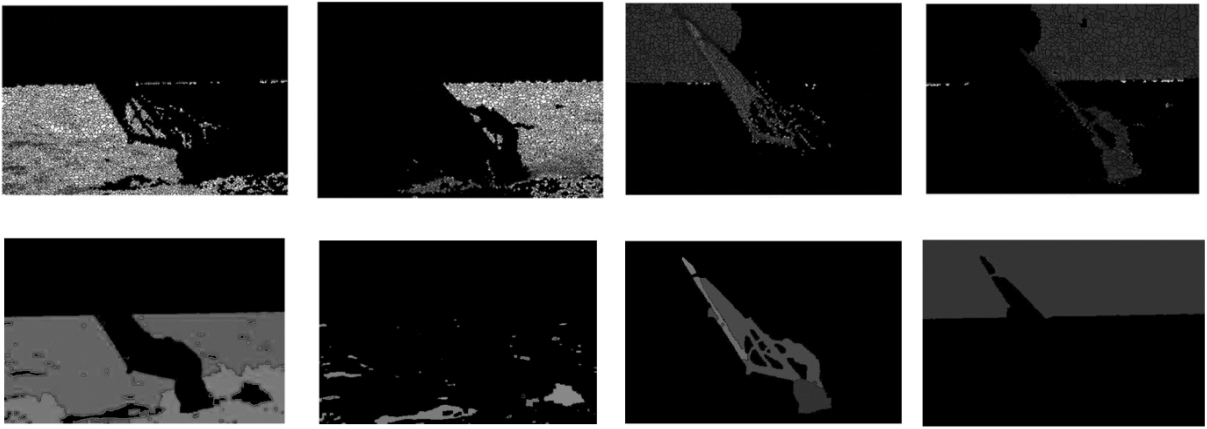

Figure 8. Comparison of our method and HS method [18] on the same image BSD – 62096. First row: main image, second row: The outcome of four steps (HS method), Third row: The outcome of four steps (out method)

### 4.3. Visual Evaluation

To evaluate our method visually, in this section, we show the results of segmentation by the proposed method on some images from BSD300. Figures 9, 10 and 11 show these results.

One of the most important advantages of proposed method is to detect the small conceptual regions as individual segments. For example, in figure 11, image 253036, our method detects all animals despite of their small size relative to image size.

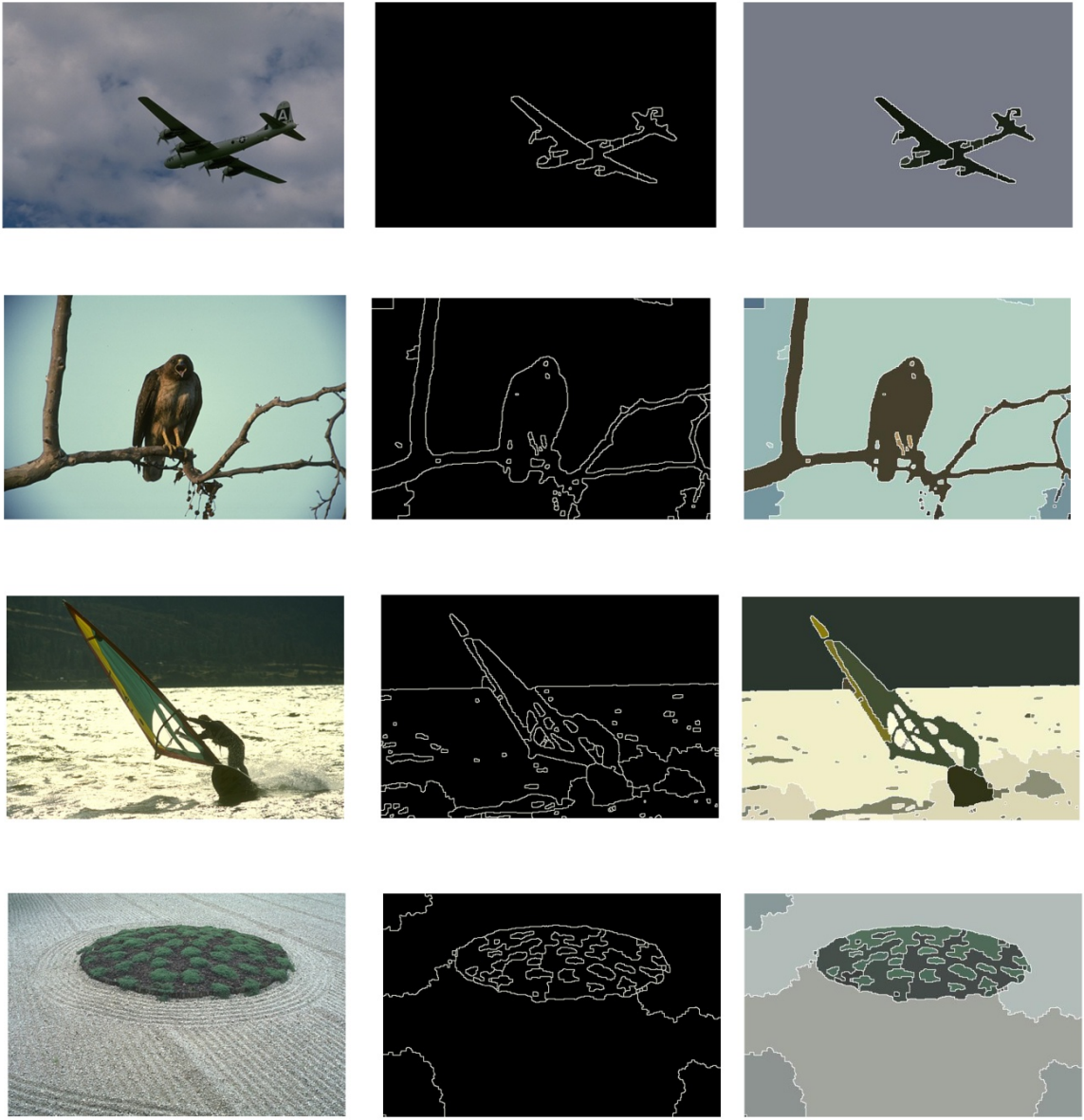

Figure 9 The results of segmentation by the proposed method on some images from BSD300 (From top: images 3096, 42049, 62069 and 86016). Left column: main images, Middle column: boundaries of final segments, Right column: final segments in average color of their pixels.

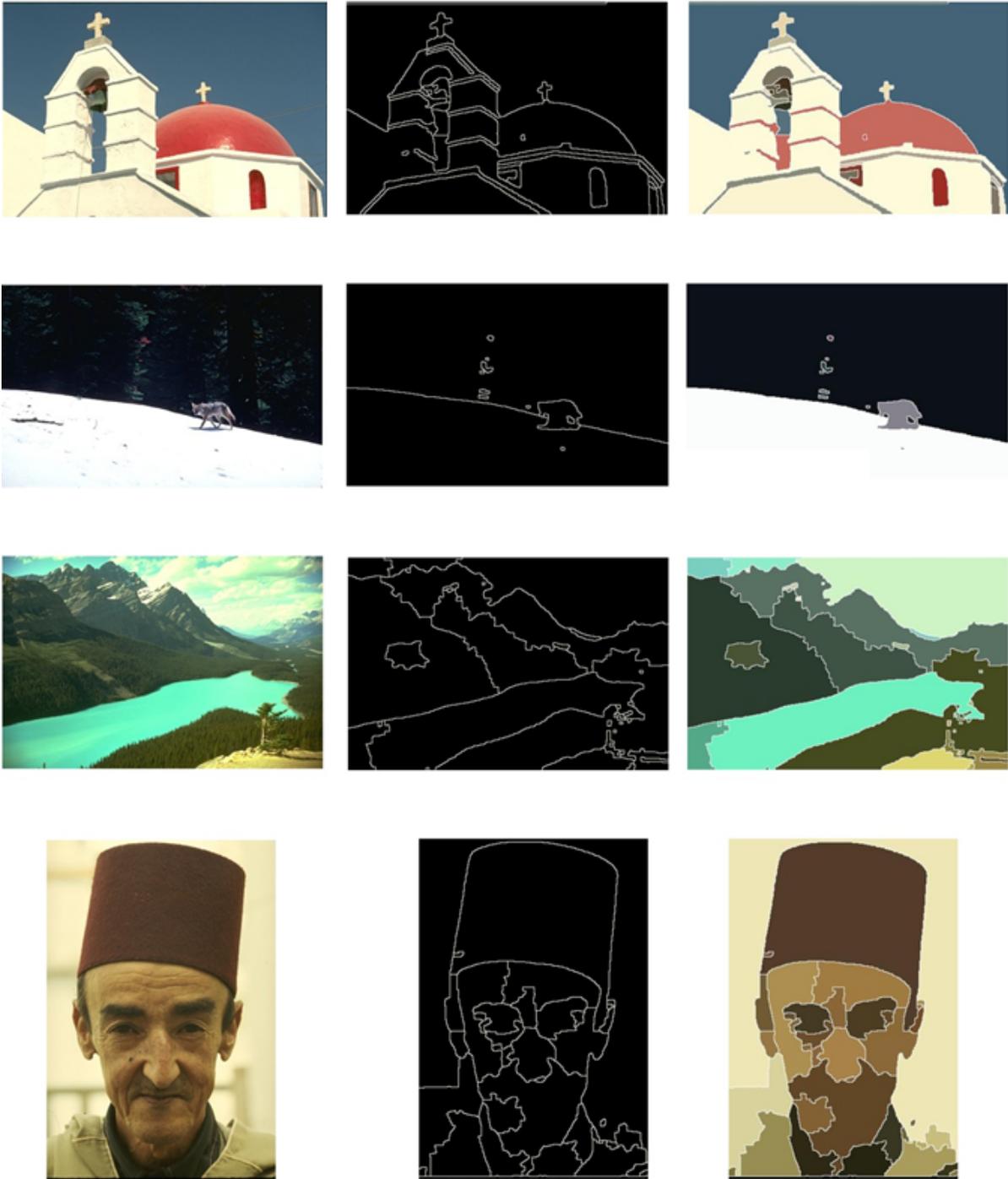

Figure 10 The results of segmentation by the proposed method on some images from BSD300 (From top: images 118035, 167062, 176035 and 189080). Left column: main images, Middle column: boundaries of final segments, Right column: final segments in average color of their pixels.

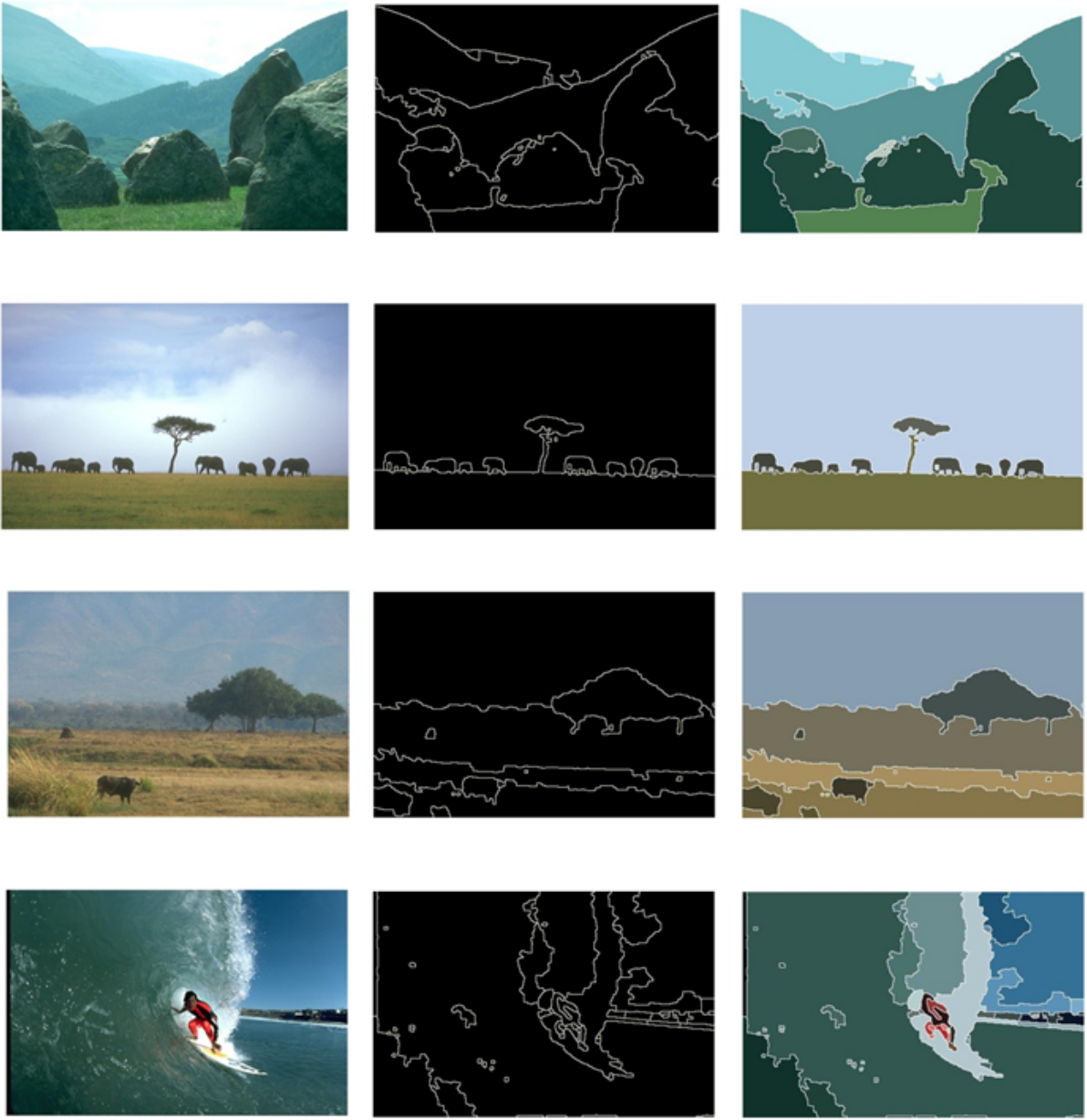

Figure 11 The results of segmentation by the proposed method on some images from BSD300 (From top: images 241004, 253036, 296007 and 300091). Left column: main images, Middle column: boundaries of final segments, Right column: final segments in average color of their pixels.

### 4.4. Weakness

Since the proposed method is based on dividing-merging approach, it's possible to merge two neighbor similar regions (that each one is belong to different segment) together. For example, consider the detected final segments by the proposed method for image 296050 from BSD300 in figure 12. An optimal segmentation algorithm is supposed to detect two elephants in different segments, while in the proposed method, these two segments is merged together.

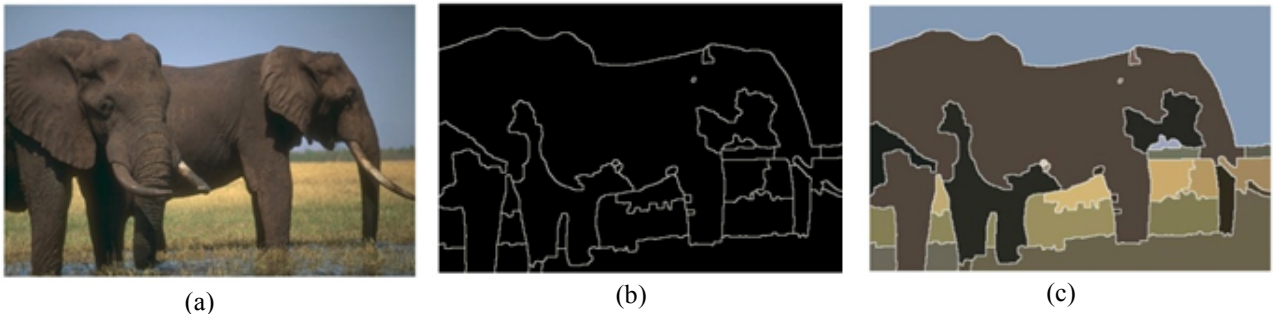

(a)            (b)            (c)

Figure 12 The results of segmentation by the proposed method on image 296050 from BSD300. (a): main image, (b): boundaries of final segments, (c): final segments in average color of their pixels.

**Conclusions**

In this paper a novel algorithm for image segmentation is proposed that is inspired from a natural behavior of human groups that are spread through a topographic plane and willing to have a social life. This algorithm has 4 steps: 1) Watershed algorithm: an edge detection method is used for determining primitive image segments. 2) Co-evolution process: the primitive segments make relation together based on their relation weights which point to similarity and proximity, then primitive segments which have high relation weights joint together and form the primitive tribes. 3) Immigration-Deportation process: the primitive segments which have not been became member of a tribe; immigrate to fittest tribe using a iterative process. 4) Emerging process: the resulted tribes which are similar highly combined together. These steps are independent from each other; therefore, each step could be changed autonomously to increase the overall performance of algorithm for specific application. Also, other features could be used instead of proposed features, for improving segmentation precision for specific application.

Experimental results show that proposed method produced continuous contours and uncomplicated boundaries. Also, the quality and quantity of segments could be controlled using parameters $\theta_p$ and $r$. In comparison with other methods, experimental results show that proposed method could be used as a novel segmentation method without using any extra information about images or segments. Also, experimental results show that in the step four, it is possible to merge two similar and close tribes together, while they are member of different final segments.

Also, because of independency of the primitive segments in the step two and independency of primitive segments and tribes in step three, it is possible to implement our algorithm parallel and improve the run time of our method. Future work will develop in three directions, first investigate other images features. Second improved proposed method for a specific application and finally using learning technique for setting parameters.

Using local information in step 2 and overall information of image in step 4 simultaneously, causes to our segmentation method be based on local and overall structures of image. Also using unsupervised learning functions for evaluating of each iteration of all four steps result in higher overall efficiency of our method.